# Autonomous Robot Swarms for Off-World Construction and Resource Mining

Jekan Thangavelautham[1]

*Space and Terrestrial Robotic Exploration (SpaceTREx) Laboratory,*
*Aerospace and Mechanical Engineering Department, University of Arizona, Tucson, US, 85721.*

Kickstarting the space economy requires identification of critical resources that can lower the cost of space transport, sustain logistic bases and communication relay networks between major nodes in the network. One important challenge with this space-economy is ensuring the low-cost transport of raw materials from one gravity-well to another. The escape delta-v of 11.2 km/s from Earth makes this proposition very expensive. Transporting materials from the Moon takes 2.4 km/s and from Mars 5.0 km/s. Based on these factors, the Moon and Mars have the potential to export material into this space economy. Water has been identified as a critical resource both to sustain human-life but also for use in propulsion, attitude-control, power, thermal storage and radiation protection systems. Water may be obtained off-world through In-Situ Resource Utilization (ISRU) in the course of human or robotic space exploration. There is also important need for construction materials such as aluminum, iron/steel, and titanium. Based upon these important findings, we have developed an energy model to determine the feasibility of developing a mining base on the Moon and Mars. These mining base mine and principally exports water, aluminum, titanium and steel. The moon has significant reserves of water known to exists at the permanently shadowed crater regions and there are significant sources of titanium, aluminum and iron throughout the Moon's surface. Mars also has significant quantities of water in the form of hydrates, in addition to reserves of iron, titanium and aluminum

Our designs for a mining base utilize renewable energy sources namely photovoltaics and solar-thermal concentrators to provide power to construct the base, keep it operational and export water and other resources using a Mass Driver. Using the energy model developed, we will determine the energy per Earth-day to export 100 tons each of water, titanium, aluminum and low-grade steel into escape velocity of the Moon and Mars. We perform a detailed comparison of the energy required for construction of similar bases on the Moon and Mars, in addition to the operating energy required for regolith excavation, processing, refining and finally transport off-the-body. In this process, we consider multiple critical technologies including use of humans predominantly to construct and operate the base and alternately the use of robot teams. In addition, we also consider the use of additive manufacturing to print a base out of local materials or use of traditional building techniques. Our comparative study finds that an equivalent Martian base requires twice as much energy for construction than a lunar base, this is to enable the base to withstand the higher gravity. This also accounts for the energy required to process the local raw material into construction feedstock. A Martian base requires significantly more energy for day to day operations due to the higher gravity, requiring 2.4-folds more energy, primarily for operating the mass-driver to export the 400-tons of export material per Earth day. More energy is needed on Mars for material extraction and for transport than the Moon, this is despite the fact that Mars gets 40% of the solar insolation of Earth. Transportation of these export resources from the Earth-Moon Lagrange points to Mars is estimated to be possible using very-low energy methods. The use of an all-robot base also minimizes the challenges of human adaptation to the low-gravity environment. These factors show a compelling reason for utilizing the Moon as a resource export economy first.

## I. Introduction

Permanent human presence beyond Low Earth Orbit (LEO) requires development of a space economy, where materials and services are readily available to sustain human day to day life, work and travel. This will require development and sustainability of critical infrastructure including communication relay networks, transport networks, fuel depots, repair and spacecraft assembly stations, storage depots, trading stations, entertainment stations, in addition to habitats for human workers to live, work and travel. The required construction material and resources need to be produced off-world to enable long-term sustainability. Transporting material off Earth is expensive, requiring a delta-v of more than 11.2 km/s. In comparison, transporting materials from the Moon requires 2.4 km/s and Mars 5.0 km/s.

---

[1] Assistant Professor, Aerospace and Mechanical Engineering, University of Arizona.

In this work, we compare the development and operation of identical resource mining bases on the Moon and Mars. Our comparison is intended to determine the energy required for construction and operation of these bases and technologies that can have game-changing impact or facilitate capabilities. At the current time, it is not clear if Mars or the Moon has a clear advantage in being able to export resources into this space economy for low-cost. The Moon is nearby Earth, has low-gravity, is less hospitable and water sources are mainly limited to the poles. Mars is nearly 8-months away, has higher gravity, most resembles Earth compared to other neighboring bodies and has resources widely dispersed on its surface.

Past surface missions to the Moon have confirmed that the Moon is nearly identical in composition to Earth (except for water). There is a diverse group of construction materials ranging from silica sand, iron, titanium, aluminum and copper available to mine. These construction material in turn can be exported into this 'space economy' to enable construction of the various infrastructure instead of launching all that material from Earth. Importantly, the relative delta-v required of transporting material off the Moon into orbit around Earth, the Lagrange points or on an escape trajectory to Mars is still less than the energy required to leave Earth's surface and enter Low Earth Orbit.

Importantly, water has also been found on the Moon but in relatively hard to reach locations. Very large deposits of water-ice are estimated to be buried as ice in the Permanently Shadowed Regions (PSRs) of the Moon. The water is unlikely to be pure and will likely have dissolved minerals including carbonates and sulphates. Sulphates in particular are of concern because they can easily poison high-efficiency electrolyzers that split water into hydrogen and oxygen. Another challenge is that the PSRs are some of the coldest, darkest places in the solar-system reaching temperature of -150 $^o$C and lower. This will likely require efficient methods to extract water ice and transport them about a 1,000 km to warmer environments (mid-latitude) for further processing and export. For this we consider use of Maglev trains that exploit super-conductivity (thanks to the cryogenic temperatures of the lunar-south pole) to levitate and transport large payloads of water ice for minimal energy. Although the operational energy required maybe minimal, the energy needed to construct this infrastructure is significant as will be shown later in this paper.

Water is as a critical 'gateway' resource that is required for refining most metals including iron, titanium, aluminum and copper. In addition, it is essential to sustain human-life and a near-universal chemical solvent. It has important applications for use in propulsion, attitude-control, power, thermal storage and radiation protection systems. Thus, water is a critical resource because of its multi-functionality. Without water, we will need to rely on unfamiliar, higher-risk 'dry' processes to extract most of the construction material known which may set us back decades.

Mars is more hospitable than the Moon with a gravity of 3.7 m/s$^2$. It has a surface environment resembling the dry desert climates of Earth but without known traces of life or a dense atmosphere. Mars has a very thin $CO_2$ atmosphere and gets 40% of the solar-insolation reaching Earth or the Moon. Water-ice has been discovered at the poles and in the form of mineral hydrates widely dispersed on the surface. Other minerals include sources of iron, titanium, copper and aluminum are widely found. Overall, the Martian crust also resembles parts of Earth, though with significant iron-oxide deposits. Even with the thin atmosphere, Mars provides a natural buffer from extreme cold and heat unlike the lunar surface. Unlike the Moon, there are multiple sources of water that can be mined on Mars and this does not require having to reach and mine from the poles.

These vastly differing but inhospitable conditions on the surface of Mars and the Moon puts into question how we might tackle the problem of resource extraction, processing and export. As hinted earlier, it is water extraction and processing that is required first to then develop extraction and refining of metals including iron, titanium, aluminum and copper. Leading aerospace companies such as United Launch Alliance (ULA), a collaboration between Boeing and Lockheed Martin is willing to buy water from any entity in space (in Low Earth Orbit, Geostationary Orbit and Lunar Orbit). ULA has put forth a plan called "CisLunar 1000" (Fig. 1) [15] an architecture that foresees nearly 1,000 people working and living in cis-lunar space and the development of communication relays [13-14], service depots, and refueling stations at strategic locations between Earth, Moon and Mars. Our proposed Lunar and Martian base design efforts is compatible with the proposed CisLunar 1000 architecture by ULA.

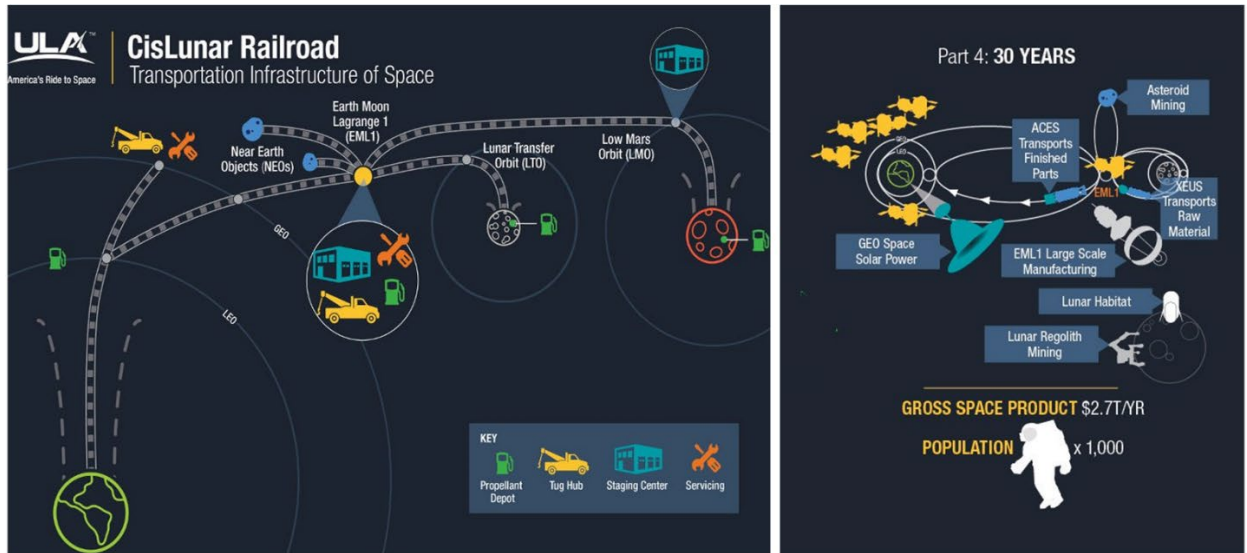

**Fig. 1. ULA's CisLunar 1000 concept envisions propellant depots, towing centers, staging centers and servicing centers located at strategic location between Earth, the Moon and Mars [15].**

In this paper, we present energy models to determine the technological feasibility of developing a Lunar and Martian base that mines and exports water, steel, titanium and aluminum on a gravity escape trajectory so that a company such as ULA can buy it for its interplanetary construction, transport and service needs. This work is a continuation of our earlier work to develop a Mars and Lunar Robotic Mining Base to export water [16, 35] and a major update to earlier effort to design a pilot ISRU base [5, 6]. Here our focus is comparing two nearly identical functional base designs to determine the differing energy costs of operating on the Moon and Mars.

Our designs for a mining base (Fig. 3) utilize renewable sources of energy from the sun namely photovoltaics and solar-thermal concentrators to provide power to construct the base and keep it operational. This includes power to transport the water (export) at gravity escape velocities using a Mass Driver (electrodynamic railgun) powered using renewable energy. In addition, on the Moon, the base power-supply will run a 1,000 km superconducting maglev rail system to transport raw, unpurified water ice from the south pole to the main mining base located in the mid-latitude region. On Mars, such a rail system will be unnecessary as there will be nearby open pit mines to extract water and all the other metals considered.

Using the energy model developed, we determined that a Mars base requires 2.3 folds more energy to operate than a Moon base to export 400-tons of export material including 100 tons of water, low-grade steel, titanium and aluminum a day. Nearly 77% of the energy needed for operations goes to extracting, transporting and refining the raw metals, while 19% of the energy goes for powering the mass-driver to eject export material into a lunar-escape velocity. Only 3.9 % of the energy is needed for transporting and processing the water ice, while the remaining 0.4% of the energy would be consumed if needed by human workers and their facilities. In comparison, for a Mars mining base, 64% of energy goes to resource extraction, transport and refining, while 34% of the energy goes for powering the mass-driver. Only 1.7% of the energy is required for water extraction and refining, while only 0.15% of the energy is needed by the human workers.

The key roadblock is the total energy required for constructing a mining base and the required supporting infrastructure. Our studies found the key to keeping the mining base simple and efficient is make it a robotic base constructed using additive manufacturing technology [1-6]. Teams of robots (consisting of 300 infrastructure robots with a mass of 120 kg each) would be used to construct the entire base using locally available resources and operate the base on a daily basis. This could result in a 24-fold energy savings on the Moon nearly identical to the 24-fold saving on Mars. Furthermore, the base can be built nearly 24-times faster using robotics and 3D printing. We estimate even greater savings are possible if we are to consider mining and refining an expanded suite of metals. The biggest road-block remains building the rail infrastructure that spans 1,000 km and that would require 35-folds more energy than construction of the main mining base.

If humans were in the loop, human energy consumption overtake nearly-all other energy needs including excavation and surface transport of resources. The only consumers that take up more energy is material refining and transport off the lunar surface. 3D printing and the 'human footprint' is significant when it comes to base construction. The human facilities significantly increase the complexity and negatively impact the overall feasibility of getting a mining base up and running. This shows that automation and robotics is the key to making such a base technologically feasible. The base would benefit from being automated as it would be located in remote off-world locations with daily tasks that are 'dull', 'dangerous' and 'dirty' and so again this is ideal for robotics.

In the following section we present related work, followed by presentation of the base architecture and a representative energy model for both construction and operation of the base in Section III. In Section 1V, we present simulation results analyzing construction, operation and use of humans on base. In Section V, we present Conclusions and plans for future work.

## II. Related Work

Initial attempts to design and build an off-world base has relied on carrying significant infrastructure, equipment and supplies from Earth. This includes the "Mars Base Camp" architecture from Lockheed Martin [26]. These base designs are human-centric and ignores use of robots entirely. Such mission requires careful design and preplanning to support an exploratory goal. Another is a concept for Mars Polar Research [27]. Here the plan is to build a human base that access the Mars polar region for long term research and data analysis. This is akin to polar research facilities in the Antarctic. The infrastructure, resources and operational supplies all need to be shipped in to sustain the base. Such attempts could be the seed to developing a long-term, self-sustaining base. A self-sustaining base need to produce goods or services to support a community and without total dependence of an external entity. The goods or services generated enables the base to buy or trade for goods and services it doesn't have and maintain a community. Mining towns in the "Old West" are good examples of a self-sustaining base. The mining town produces one or two products of value that generates revenue to help uplift its entire population. This is in contrast to a self-sufficient base where all the requires good and services are produced in-house within the town or camp.

There have been ambitious attempts at building simulation facilities on Earth to determine how humans can adapt and operate in these enclosed and isolated bases and attempt to achieve self-sufficiency. This includes past ambitious efforts such as Biosphere 2 (Fig. 2 left) located in the desert of Southern Arizona and that aimed to have a fully self-sufficient base with miniature support ecosystem to help feed and keep alive a team of human residents for 2 years [24-25]. The experiments produced mixed results, with the ambitious goals of self-sufficiency never being achieved. These results show increased complexity involved in having a base that support humans and living organisms. Similar experiments are being performed at a smaller scale by the Chinese Space Agency (CNSA) by utilizing an enclosed mock Martian base in the inhospitable Gobi Desert [28]. Another ambitious effort is being pursued by the United Arab Emirates Space Agency with the development of a 136-million-dollar "Mars Science City" (Fig. 2 right) being built in the Arabian desert [29].

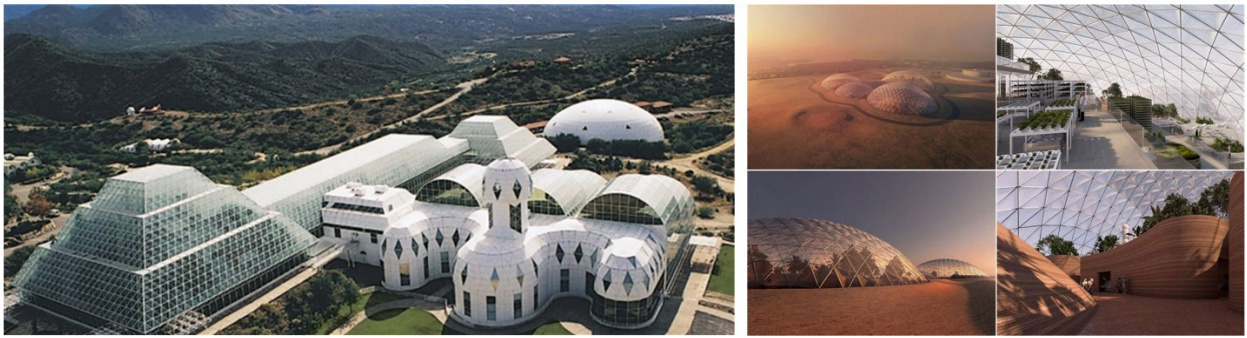

**Fig. 2. (Left) Biosphere 2 located in Southern Arizona is a 3.14-acre indoor (previously sealed) facility designed to operate as a self-sufficient base housing 8 humans for 2 years. Two experiments were conducted in 1991 and 1993, leading to mixed results with self-sufficiency never being achieved [24-25]. (Right) UAE Mars Science City Concept.**

The multitude of challenges of sustaining humans and human-centric facilities in space as seen from experiments over the past 20 years has led to consideration of alternate pathways [31]. Exploring and operating in space is both challenging and complex. The off-world sites from which critical resources need to be extracted are inhospitable, the day to day tasks dull, dirty and dangerous making it well suited for robotics. Robotics has meanwhile significantly advanced over the past 20 years, with rapid progress being made in robot dexterity, vision, planning, control and construction. There is a growing list of tasks where robotic systems are human competitive (at the levels of humans) and instances where they exceed even leading human experts [2]. There now exists entire factories that are fully automated with robots. Robots are also making inroads with self-driving cars, trucks and aircraft. Robots have also started to emerge as helpers in the home and workplace taking on dull, dangerous and dirty tasks.

One of the main enabling technologies for an off-world base to become reality is through automation of construction tasks. To date most automated construction methods build components but are yet to build whole houses or buildings. One technology called Contour Crafting [7, 32-33] can be operated on a gantry or large mobile robot and can craft the walls of a building, a layer at a time. Contour crafting uses wet cement like paste. Other approaches have suggested using solar thermal printing to melt sand and lose rock into glass [30, 34].

### III. Off-World Mining Base

The proposed layout of the Robotic Mining Base (RMB) is shown in Fig. 3 and covers nearly 2 square kilometers. The mining base would be located at the base of a crater to exploit use of natural incline (slope) for the Mass Driver [11]. Key facilities on the base are interconnected by roadways constructed out of heat-fused silica (as a replacement to concrete). The command and control facilities of the base consist of a 150-meter high control tower to monitor/verify all operations on the base. The command tower will also be a localization beacon and tracking system for the fleet of autonomous robots. In addition, the base is equipped with a communication ground station to communicate directly with Earth through the Deep Space Network (DSN) or equivalent. On the Moon, raw materials such as water ice will be transported daily by superconductive Maglev rail from the lunar south pole. While on Mars, the raw materials will be brought by a fleet of robotic trucks along the paved roadway.

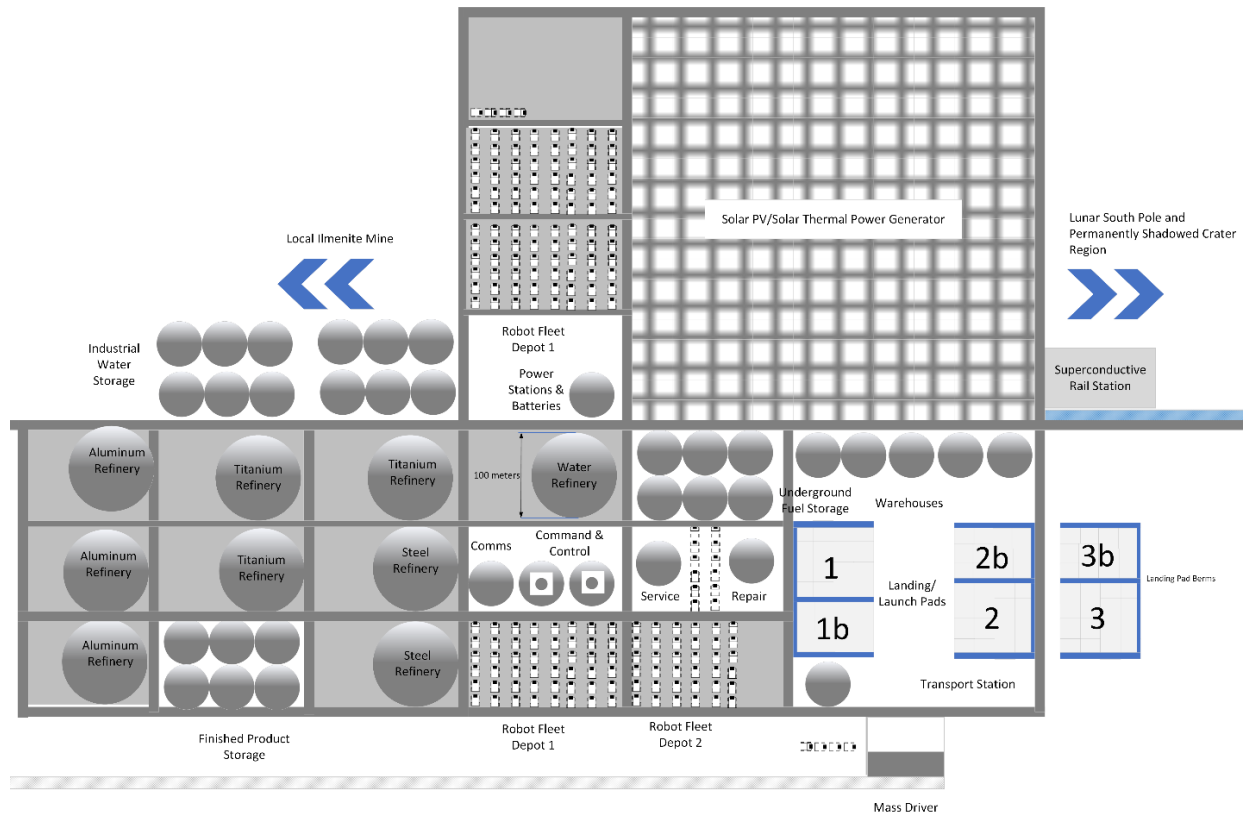

**Fig. 3. Layout of a Lunar Robotic Mining Base. It occupies over 2 sq. km and would be situated at the base of a crater with known resource deposits. In this concept, there are no humans occupying the base.**

Major facilities on the base including the refinery, service and repair center, command and control buildings, Maglev station and even the warehouses are all 3D-printed spherical domes. The buildings are spherical domes to maximize internal volume with very minimal construction material. The domes would be constructed in one piece, include large glass windows, with an airtight fused silica glass layer in between to optionally maintain a 100 KPa, nitrogen/oxygen atmosphere inside.

A significant portion of the base is covered in solar-thermal and photovoltaic panels to harvest energy from the sun. Furthermore, solar-reflectors will be located on the crater rim to increase net-sunlight beamed to the base, in-addition to raising the temperature of the base surroundings. The base includes a hardened depot to house the 300 robotic vehicles when not in use, in-addition to service and repair facilities. The base houses several refineries to process the regolith into steel, aluminum, titanium and liquid water. An underground facility will be used to store both water and rocket propellant to refuel incoming rockets.

The human habitat and human facilities will be separated from the central operations area of the base by the Solar PV and solar-thermal generators. This is to minimize dust and sand being churned by moving vehicles entering and exiting the base from entering the human habitat regions. In addition, it will keep the human habitat sectors well-away from the robotic vehicle traffic.

There are up to three modes of transport from the base. This includes roadways interlinking, (1) the base to the open pit mine site where regolith containing metals are located, (2) optional superconductive rail reaching the lunar south pole and Permanently Shadowed Regions (PSRs) and (3) a service road for the Mass Driver. The third mode of transport to and from the base is using rockets that can vertically land and take-off from one of six landing pads at the edge of the base. This provides quick access to the base in case of critical maintenance and for transporting repair equipment and other seed-resources to maintain continual operations of the base. Finally, the fourth form of transport is the mass-driver that will propel water, processed titanium, low-carbon steel and aluminum housed in containers to a velocity of 2.4 km/s on the Moon and to a velocity of 5.0 km/s on Mars.

In our simulation studies, we consider four scenarios: Base constructed using (1) 3D printing of fused silica sand, with steel rebar support [7-10] and (2) Steel structure with internal silica-sand blocks. Secondly, we consider the base to be (a) fully autonomous run using up to 300 mobile robots (Fig. 4) and (b) base run by a human team of 300 workers. The human occupied base as would be expected is much more complex than the robotic base and includes additional buildings shaded in gray (Table 1). Importantly for the human occupied base, there will be $O_2$ electrolyzers and recyclers. While every effort will be done to recycle $O_2$, some will be lost due to everyday activity at the base and new $O_2$ will need to be generated from electrolysis of $H_2O$ on the Moon or electrolysis of $CO_2$ on Mars.

The base will also contain two large domes as large as the one of the nine refineries housing services to take care of the human occupants, including health, banking, pharmaceutical, shopping, restaurants, rest and relaxation centers. The human occupants will live in individual dome-shaped housing with a minimal floor area 120 m$^2$ or 1,200 sq. feet. The housing areas will be linked via enclosed/pressurized above ground and underground walkways (tubes) spanning 10 km and connecting all the buildings. Several parallel tubes are in place to enable redundant and secure access to key facilities. In addition, there are parallel tunnels leading straight from the living quarters to the launch pad for quick evacuation in case of fire or a major onsite accident.

The Off-World Mining Base contain the following infrastructure. Rows shaded in grey are included if the base is to be occupied and operated by human workers:

Table 1. Off-World Mining Base Structures

| Structure | Quantity | Dimensions |
|---|---|---|
| Road Network [t] | 1 | Length, $L_t$ = 12,000 m, Width, $W_t$ = 8 m, Depth, $d_t$ = 0.2 m, Steel Ratio, $S_t$=0.05 |
| H$_2$O and Metal Refineries [r] | 9 | Dome, Outer Radius, $R_{or}$ = 50 m, Inner Radius, $R_{ir}$ = 49.6 m, Base Depth, $D_r$=0.3 m, Steel Ratio, $S_r$=0.1 |
| Comms. Command, Control, Service and Repair, Power Ctrl [s] | 6 | Dome, Outer Radius, $R_{os}$ = 25 m, Inner Radius, $R_{is}$ = 24.6 m, Base Depth, $D_s$=0.3 m, Steel Ratio, $S_s$=0.1 |
| Control Tower [CC] | 1 | Pyramidal Tower, Height, $H_{cc}$=150, Base Length, $L_{1cc}$=25, Top Length, $L_{2cc}$=5 |
| | | Steel Ratio, Scc= 0.1 |
| Warehouses [w] | 21 | Dome<br>Outer Radius, $R_{ow}$ = 25 m<br>Inner Radius, $R_{iw}$ = 24.8 m<br>Base Depth, $D_w$=0.3 m<br>Steel Ratio, $S_w$=0.1 |
| Fuel Storage [f] | 6 | Cylinder<br>Radius, $r_f$ = 25 m<br>Wall Thickness, $D_f$=0.1 m<br>Wall Height, $h_f = 2\ m$<br>Steel Ratio, $S_f$=0.05 |
| Small Landing Pad [p] | 3 | Pad and Blast Walls<br>Length, $L_p$ = 100 m<br>Width, $W_p$ = 50 m<br>Depth, $D_p$ = 0.2 m<br>Wall Height, $H_p$ 0.5 m<br>Steel Ratio, $S_p$=0.05 |
| Large Landing Pad [p] | 3 | Pad and Blast Walls<br>Length, $L_p$ = 100 m<br>Width, $W_p$ = 100 m<br>Depth, $D_p$ = 0.2 m<br>Wall Height, $H_p$=0.5 m<br>Steel Ratio, $S_p$=0.05 |
| Power Generation Pad [g] | 1 | Pad<br>Length, $L_g$=1000 m<br>Width, $W_g$= 1000 m<br>Depth, $D_g$=0.1 m<br>Steel Ratio, $S_g$=0.05 |
| Mass Driver [m] | 1 | Cylinder (Pair) + Support<br>Inner Radius, $R_m$ = 3 m<br>Length, $L_m$ = 10,000 m<br>Thickness, $T_m$ = 0.5 m<br>Slope, $\alpha_m$ = 5°<br>Post Height= $H_m$= 5 m<br>Post Width = $W_m$ = 1 m<br>Post Spacing = $\varepsilon_m$ = 20 m<br>Post Steel Ratio, $S_m$=0.05 |
| Superconductive Rail [SR] | 1 | Cylinder (Pair) + Support<br>Inner Radius, $R_{SR}$ = 3 m<br>Length, $L_{SR}$ = 1000 km<br>Depth-magnet, $D_{SR}$=0.1m<br>Thickness, $T_{SR}$ = 0.5 m<br>Post Height= $H_{SR}$= 5 m<br>Post Width = $W_{SR}$ = 1 m<br>Post Spacing = $\varepsilon_{SR}$ = 20 m<br>Post Steel Ratio, $S_{SR}$=0.05 |
| $O_2$ from $H_2O$ Extract* [O] | 18 | Cylinder<br>Radius, $r_o$ = 50 m<br>Wall Thickness, $D_o$=0.2 m<br>Wall Height, $h_o$=3.0 m<br>Steel Ratio, $S_o$=0.2 |
| Human Services Centers* [SC] | 6 | Dome<br>Outer Radius, $R_{or}$ = 50 m<br>Inner Radius, $R_{ir}$ = 49.6 m<br>Base Depth, $D_r$=0.3 m |
| | | Steel Ratio, $S_r$=0.1 |
| Human Habitat* [h] | 300 | Dome<br>Outer Radius, $R_{oh}$ = 6 m<br>Inner Radius, $R_{ih}$ = 5.9 m<br>Base Depth, $D_h$=0.1 m<br>Steel Ratio, $S_h$=0.2 |
| Human Walkways (Service Tubes) [ST] | 1 | Cylinders<br>Length, $L_{ST}$ = 10,000 m,<br>Outer Radius, $R_{oST}$ = 1.1 m<br>Inner Radius, $R_{iST}$ = 1.0 m<br>Steel Ratio, $S_{ST}$=0.1 |

The equations for volume of each structure is given below:

$$V_t = L_t W_t D_t \tag{1}$$

$$V_r = \frac{2}{3}\pi(R_{or}^3 - R_{ir}^3) + \pi D_r R_{or}^2 \tag{2}$$

$$V_s = \frac{2}{3}\pi(R_{os}^3 - R_{is}^3) + \pi D_s R_{os}^2 \tag{3}$$

$$V_{CC} = H_{cc}L_{2cc}^2 + H_{cc}(L_{1CC} - L_{2CC})L_{2CC} \tag{4}$$

$$V_w = \frac{2}{3}\pi(R_{ow}^3 - R_{iw}^3) + \pi D_w R_{ow}^2 \tag{5}$$

$$V_f = 2\pi r_f^2 D_f + 2\pi r_f h_f D_f \tag{6}$$

$$V_p = L_p W_p D_p + 2L_p H_p D_p + W_p H_p D_p \tag{7}$$

$$V_g = L_g W_g D_g \tag{8}$$

$$V_O = 2\pi r_o^2 D_O + 2\pi r_o h_o D_O \tag{9}$$

$$V_{SC} = \frac{2}{3}\pi(R_{oSC}^3 - R_{iSC}^3) + \pi D_{SC} R_{oSC}^2 \tag{10}$$

$$V_h = \frac{2}{3}\pi(R_{oh}^3 - R_{ih}^3) + \pi D_h R_{oh}^2 \tag{11}$$

$$V_m = 2\pi L_m (R_m + T_m)^2 - 2\pi L_m (R_m)^2 + \frac{L_m}{\varepsilon_m} 2H_m W_m^2 \tag{12}$$

$$V_{SR} = 2\pi L_{SR}(R_{SR} + T_{SR})^2 - 2\pi L_{SR}(R_{SR})^2 + D_{SR}W_{SR}L_{SR} + \frac{L_{SR}}{\varepsilon_{SR}} 2H_{SR}W_{SR}^2 \tag{13}$$

$$V_{ST} = \pi(R_{oST}^2 - R_{iST}^2)L_{ST} \tag{14}$$

A. **Energy for Silica Sand 3D Printing + Reinforcement**

The energy needed for silica sand 3D printing involves first raising the temperature of the sand to $\Delta T_{sand}$ =1,973 °K followed by melting the sand into a liquid binder. The heat capacity of sand (quartz) is, $C_{p\ sand}$= 0.830 kJ/(kg K). The heat of fusion to melt sand (quartz) is $H_{melt}$=156 kJ/kg. Therefore, the total energy needed to melt and fuse the sand in-situ is the following:

$$E_{total\ sand} = E_{heat\ sand} + E_{melt\ sand} + E_{trans\ sand} \tag{15}$$

$$E_{total\ sand} = \rho_{sand}V(C_{p\ sand}\Delta T + H_{melt\ sand} + gd) \tag{16}$$

The density of quartz sand is $\rho_{sand} = 1500\ kg/m^3$ and $g$ is the local acceleration due to gravity and $d$ is distance in meters. On the Moon, g=1.62 m/s². Unless if the sand is transported long distances in the order of 1000s of meters, the energy required simplifies to the following:

$$E_{total\ sand} \approx E_{heat\ sand} + E_{melt\ sand} \tag{17}$$

For steel-bar reinforced structures, the total energy is the following:

$$E_{total\ reinf} = E_{total\ sand} + E_{total\ steel} \tag{18}$$

$$E_{total\ reinf} = \rho_{sand}(1-S)V(C_{p\ sand}\Delta T_{sand} + H_{m\ sand}) + SV\rho_{steel}(C_{p\ steel}\Delta T_{steel} + H_{m\ steel}) \tag{19}$$

Where $S$ is the volumetric percentage of steel, $\rho_{steel} = 7{,}750\ kg/m^3$, $C_{p\ steel}$ = 0.510 kJ/(kg K) and $H_{m\ steel}$ = 25.23 MJ/kg where the steel is produced from regolith containing 1:1, Magnetite and Hematite with heat of melting 1,118 kJ/mol and 824 kJ/mol. Only 70% of Magnetite and Hematite contain Iron, with the remainder being oxygen. $\Delta T_{steel}$ =1,640 °K

### B. Steel and Sand Block Construction

For the steel and sand-block construction, the structures are all made of a steel support structure and thus the volume of the steel structure is µ=0.15 of the overall 3D printed structure. For roads, the volume of the structure is µ=0.05 of the overall 3D printed structure. Similarly, to form the sand blocks, only µ percentage volume of the sand is melted thus the energy equation is the following:

$$E_{total} = \rho_{steel}V\mu(C_{p\ steel}\Delta T + H_{m\ steel}) + \rho_{sand}V\mu(C_{p\ sand}\Delta T + H_{m\ sand}) \tag{20}$$

### C. Water Extraction Energy from Ice

We wish to extract liquid water from the ice sources found at the pole. The total energy is the following:

$$E_{total\ water} = m_{water}(C_{p\ water}\Delta T_2 + H_{melt\ ice} + C_{p\ ice}\Delta T_1) \tag{21}$$

Presuming the ice is found at -150 °C and it is warmed to a liquid at 25 °C Then $\Delta T_1$=150 and $\Delta T_1$=25. $C_{pwater}$ is the specific heat of water and is 4.200 kJ/kg°C. The specific heat of water ice, $C_{p\ ice}$ =2.108 kJ/kg°C and $H_{melt\ ice}$ is 333.55 kJ/kg and we have a total mass of 100,000 kg of water. The total energy needed to bring it to liquid water at 25 °C is 7.55 × 10⁵ MJ.

### D. Water Extraction Energy from Hydrates

We may also wish to extract water from regolith contained in the form of hydrates. This is applicable for Mars and not the Moon. We model the scenario as the regolith contain 90.6 % sand (quartz) and 9.4 % $MgCl_2 \cdot 6H_2O$. This gives us ~5 % $H_2O$ by mass. To extract water from m= 100,000 kg, from $MgCl_2 \cdot 6H_2O$ requires raising the temperature to 111 °C from a presumed ambient of 25 °C, where the heat capacity Cpmg= 0.756 kJ/(kg K) The heat of dehydration is the $E_{dehyd}$= 138 kJ/kg. Furthermore, the extracted water needs to be condensed down to 25 °C. Therefore, the total energy required is the following:

$$E_{heat} = E_{dhyd}m(9.4/5) + (111\text{-}25)\,[2C_{pH2O}\,m + \ + C_{pmg}\,(4.42/5)\,m + \ C_{psand}\,m(90.6/5)] \tag{21b}$$

Based on this estimate, it takes $E_{heat}$ = 4.37 × 10⁵ MJ to extract the water from the regolith.

### E. Excavation and Regolith Transport Energy Requirements

We wish to calculate the total excavation energy required to produce 100,000 kg of water during a 24 hour period. First, we calculate the transport energy required to move lunar regolith from a permanently shadowed crater containing water ice 90% by mass, a distance d = 5000 m, with friction coefficient η=0.01, χ= 1.2 and is the vehicle movement ratio.

$$E_{mov\ up} = (100/90)\ mg\eta d\chi \qquad (22)$$

Plugging in the values, $E_{mov\_unp}$ = 10.7 MJ. Next we calculate the total transport energy required to transport the processed water from refinery to Mass Driver. We presume a maximum d= 1000 m, with friction coefficient η=0.01:

$$E_{mov\ p} = mg\eta d. \qquad (23)$$

For hauling the water ice $E_{mov\ p}$ = 8 MJ. Next, we calculate an estimate of the total excavation distance required to cover a flat open pit containing water ice. Knowing that the $\rho_{ice}$ = 1000 kg/m³, we determine the total volume of the input material. Following this we divide by the total area of the robot vehicle, $A_{robot}$ = 0.42 m² and account for picking up and returning with regolith. The expression is then the following:

$$E_{dig} = [\chi F_{dig} + (100/90)mg\eta]\ [(100/90)\ m/\rho_{ice}]\frac{2}{A_{robot}} \qquad (24)$$

Where χ= 1.2 is vehicle movement ratio and accounts for inefficiencies in movement. $F_{dig}$ = 3000 N is the force used to dig into the lunar surface and m= 100,000 kg and η=0.1 is the friction coefficient and is applicable for movement in rough terrain. Then $E_{dig\ p}$ = 113 MJ.

Finally, we determine the energy need to transport raw water ice from the permanently shadowed crater region to the mining base for processing using the superconductive rail. We presume d=1,000,000 m. We estimate the rail carriage mass to have a 20% mass overhead of the cargo and thus φ= 1.2 and η=0.005. Therefore, the energy need to haul the mass by rail is the following:

$$E_{train} = \varphi mg\eta d \qquad (25)$$

Therefore, the total energy required for train transport is $E_{train}$ = 972 MJ. With this the total energy required for excavation, transport of unprocessed regolith, water extraction and transport of water to Mass Driver is $E_{Exc\_transp}$ = 1.64× 10⁶ MJ.

### F. Energy Needed for extracting iron (steel) and titanium

The mining base apart from mining water will also be mining equivalent of 100 tons of low-grade steel and 100 tons of titanium and 100 tons of aluminium. Iron and titanium materials can be extracted from ilmenite (FeTiO₃) that is located in plentiful supply on the surface of the Moon and Mars. Titanium makes up, χ= 31.6% of ilmenite, while iron makes up, χ=37 %. Presuming a 70% extraction efficiency (κ) for each metal. Then the following amount of ilmenite needs to be excavated:

$$m_{ilmenite} = \frac{m_{Ti}}{\kappa \chi_{Ti}} \qquad (26)$$

The total mass of the ilmenite then is $m_{ilmenite}$ = 4.53 × 10⁵ kg. At an ilmenite deposit (mine), we presume 80% ilmenite by mass and therefore is the following:

$$E_{mov\ up} = (100/80)\ mg\eta d\chi \qquad (27)$$

$E_{mov\ up}$ for the ilmenite is the 55 MJ for every ~100 tons of Ti and Fe. The total $E_{mov\ p}$ for the titanium and iron(steel) haul is the following:

$$E_{mov\ p} = mg\eta d \tag{28}$$

Where m=200,000 kg, d= 1000 m, η=0.01 and therefore the total energy for $E_{mov\ p}$ = 3.2 MJ. The digging energy required for the ilmenite is the following:

$$E_{dig} = [\chi F_{dig} + (100/80)mg\eta]\ [(100/80)\ m/\rho_{ilmenite}]\frac{2}{A_{robot}} \tag{29}$$

Where $\rho_{ilmenite}$ = 4800 kg/m³. Then the total $E_{dig}$ is 53.5 MJ. Finally, it should be noted the ilmenite mine will be in surrounding region around the mine and hence it doesn't need to be hauled by train. Next, we consider the energy required to produce low-grade steel and iron from the raw ilmenite (Table 2). Terrestrial methods of steel and titanium production also requires plenty of water. Extracting and using clean water for these industrial processes becomes even more energy intensive as for this Lunar Mining Base it takes 16 MJ/kg of water produced. Therefore, a lower energy cost option is to use recycled water for titanium and steel production. Water can be recycled using reverse-osmosis systems, with a minimum energy requirement of 3 kWh/m³ for brackish water which is 0.0108 MJ/kg. It is more likely the water used contain undissolved particles rather than salts such as NaCl and therefore require closer to the minimal energy for reverse osmosis and passive sand filtration to occur.

**Table 2. Resource Processing Energy & Water Needs**

| Material | Energy Req. | Water Needed | Recycled Water Energy |
|---|---|---|---|
| Low-Grade Steel | 25 MJ/kg | 23 L/kg | 0.25 MJ/kg |
| Stainless Steel | 85 MJ/kg | 112 L/kg | 1.2 MJ/kg |
| Titanium | 120 MJ/kg | 190 L/kg | 2.1 MJ/kg |
| Aluminium | 138 MJ/kg | 200 L/kg | 2.1 MJ/kg |
| Water* | 16 MJ/kg | - | |

For 100 tons of low-grade steel, the $E_{process} = mE_{p\ req}$ = 2.5 × 10⁶ MJ and for titanium, $E_{process}$ = 1.21 × 10⁷ MJ. For aluminium it is $E_{process}$ = 1.38 × 10⁷ MJ. The total energy required to process 100 tons of low-grade steel, 100 tons of titanium and 100 tons of aluminium is 2.84 × 10⁷ MJ. This energy includes use and purification of recycled water using reverse-osmosis.

### G. Mass Driver Energy Needs

For the proposed mining base, a Mass Driver will be used to send the m=100,000 kg of liquid water in a container consisting of M=10,000 kg into a Moon escape velocity. The
Escape velocity of the Moon is the following:

$$v_{escape\ moon} = \sqrt{\frac{2GM_{moon}}{r_{moon}}} \tag{30}$$

Substituting for the estimated mass of the Moon, $M_{moon}$= 7.34 ×10²² kg and $r_{moon}$ = 1,731 km, the Moon escape velocity is the following, $v_{escape\_moon}$=2.44 km/s. For Mars the escape velocity, is 5.017 km/s. The total energy requirement is simply the following for the Moon as it doesn't have an atmosphere:

$$E_m = \alpha \frac{1}{2}(\varphi m)v^2 \tag{31}$$

Where $\varphi = 1.5$ and is the container and $\alpha$ is the efficiency of the Mass Driver is presumed to be 2.0. The total energy needed to export all of the resources is then $E_{m\ total}$=3.567 × 10⁶ MJ. For Mars, $E_{m\ total}$=1.51 × 10⁷ MJ.

### H. Energy Requirements for Work Crew

Here is a table of daily energy requirements for the 300-member work crew:

Table 3. Life Support Energy Consumption

| Item | Energy Required (Moon) | Energy Required (Mars) |
|---|---|---|
| Oxygen Generation | 16,676 MJ | 16,676 MJ |
| Electricity | 16,200 MJ | 10,800 MJ |
| Water Needs | 32,400 MJ | 32,410 MJ |
| Food | 10,062 MJ | 10,064 MJ |
| Total | 75,339 MJ | 69,950 MJ |

These figures are based on expected requirements for an average adult human. An average adult is expected to breathe 98.2 moles of $O_2$ a day. The energy required to recycle $O_2$ from $CO_2$ is 566 kJ/mol. Each person is estimated to consume 15 kWh per day (on the Moon) and 10 kWh per sol (on Mars) for personal use which is 16,700 MJ on the Moon and 10,800 MJ on Mars 300. The difference in electricity consumption is due to more heating required during lunar nights on the Moon. Each worker is assumed to consume 300 L in water a day. The water is recycled and presumed to consume 1 kWh/L. In terms of food, the following Table 4 shows kJ energy consumed in terms of food person. The differences in energy consumed for food and water on Mars vs. Moon is due to differing transportation energy costs. Considering 50% of food goes to waste, we presume the total requirements of food per person sol is 16.77 ×2 kJ = 33.54 kJ.

Table 4. Human Daily Food Consumption

| Food | Energy Used [kJ/kg] | Consumed [kg] | Total Energy [kJ] |
|---|---|---|---|
| Corn | 1.1 | 0.306 | 0.336 |
| Milk | 2.2 | 0.18 | 0.39 |
| Fruits & Vegetables | 4.4 | 0.864 | 3.80 |
| Eggs | 8.36 | 0.09 | 0.75 |
| Chicken | 8.8 | 0.09 | 0.79 |
| Cheese | 17.6 | 0.09 | 1.58 |
| Goat | 30.8 | 0.09 | 2.77 |
| Beef | 70.4 | 0.09 | 6.33 |
|  |  | 1.8 | 16.77 |

Based on these calculations, the total energy requirements to support the 300 human workers is 75,339 MJ/day on the Moon and 69,950 MJ/sol on Mars (Table 3).

**I. Energy Harvested for the Mining Base**

The proposed Moon and Martian mining base will be using renewable energy to power the entire facility. The Moon mining base needs to generate $1.89 \times 10^7$ MJ/day in terms of electrical and solar thermal energy and where we are counting in terms of Earth days. For this we make some simplifying assumptions. The average daytime solar insolation on Moon is 1360 W/m². Solar-thermal systems that use Carbon Nano-Tubes (CNTs) can directly convert solar energy into heat at 99% efficiency [21]. Using CNT, we consider assembly of a solar-thermal plant that capture solar heat. We presume we require generating $1.89 \times 10^7$ MJ/day of thermal power and furthermore sunlight hours last 24 hours/day and the average solar insolation at the Moon, $\chi=1360$ W/m². The total size of the solar-thermal power plant at the Moon base is the following:

$$E_{solar} = \Delta T \chi A_{solar} \qquad (32)$$

This requires a total area of 0.15 km². Next, we presume the total energy required per day is all electricity. Then equation for the total energy required is the following:

$$E_{PV} = \Delta T \chi \lambda_{PV} A_{PV} \qquad (33)$$

Where $\lambda_{PV}$ is the photovoltaic efficiency. We presume it is 45%. With these parameters, the required Area, $A_{PV}$ is 0.33 km².

The Martian base needs to generate $4.42 \times 10^7$ MJ/sol and the solar insolation on Mars is a paltry 544 W/m². We presume daylight last 8 hours/day and the average solar insolation at Mars, $\chi=544$ W/m². Using (32), it requires a total area of 2.82 km². The total area of the PV field using (33) is then 6.27 km².

## IV. Results and Discussion

Autonomous robotics and 3D printing can have game-changing impact on off-world mining base construction and operation. We developed an energy model that accounts for construction, operation and maintenance of an Off-World Mining Base. In this study, we also considered the use of 300 human workers vs. 300 infrastructure robots to operate and maintain the base.

### A. Energy Required for Construction

Our studies compared the potential options for building a Moon (Fig. 4, left) and Martian Mining Base (Fig. 4, right). This included use of humans or robots and 3D-printing or no 3D printing. Our studies find that 3D printing and robotics provides the biggest advantage. It can decrease energy consumption by 23-folds (Fig. 4) and speed-up construction by 23-folds (Fig. 6).

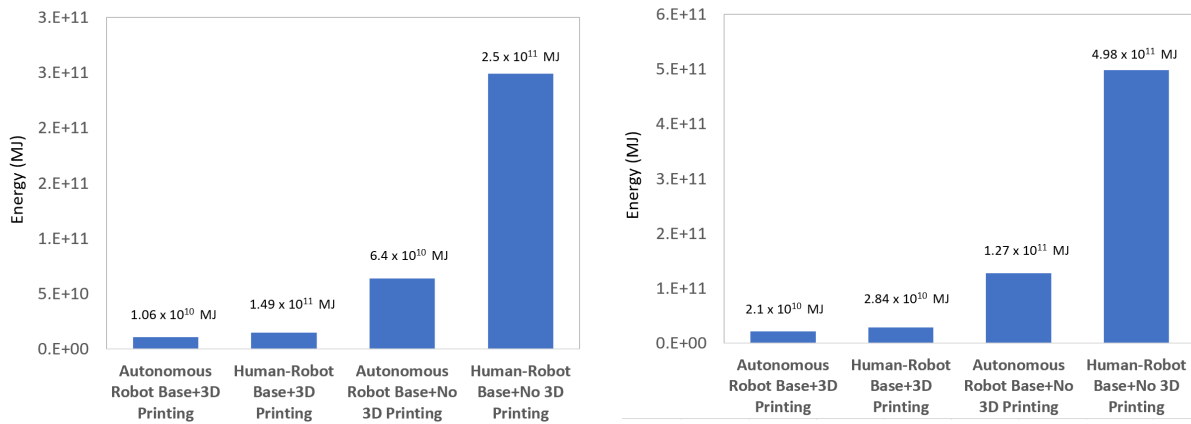

**Fig. 4. (Left) Building a Moon Mining Base using a human team and without 3D printing requires 23-folds more energy than without. Similar results are obtained if we consider a Martian Mining Base (Right). This shows robotic 3D-printing has significant potential to lower cost and make feasible off-world mining.**

Furthermore, we compare base construction to cost of other major infrastructure such as the 1,000 km magnetically levitated rail-line (Fig. 5) to haul raw ice from the lunar Permanently Shadowed Regions (PSRs) to the base.

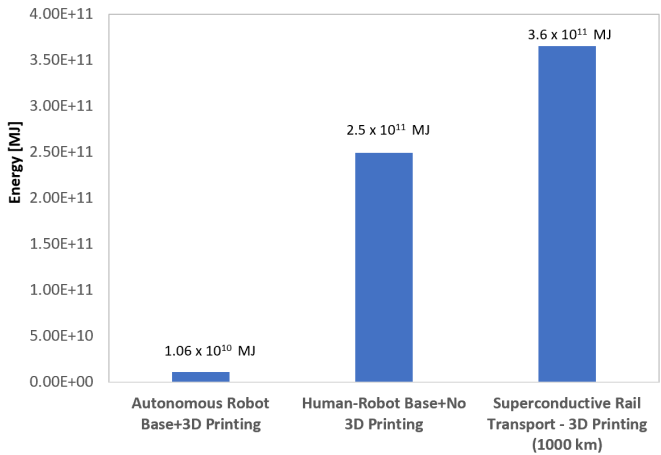

**Fig. 5. Our models show that building a 1,000 km superconductive rail system would require significantly more energy (35-folds) more than an all robotic 3D printed base and 1.5-folds more than the human occupied non-3D printed base.**

It would take nearly 35-folds more energy to build this 1,000 km rail line than the main base facility. In comparison to the human operated non-3D printed base, it would be 1.5-folds more energy. For this rail line, we only considered the 3D printing option, as the non-3D printing option would require exorbitantly high energy, making it unfeasible.

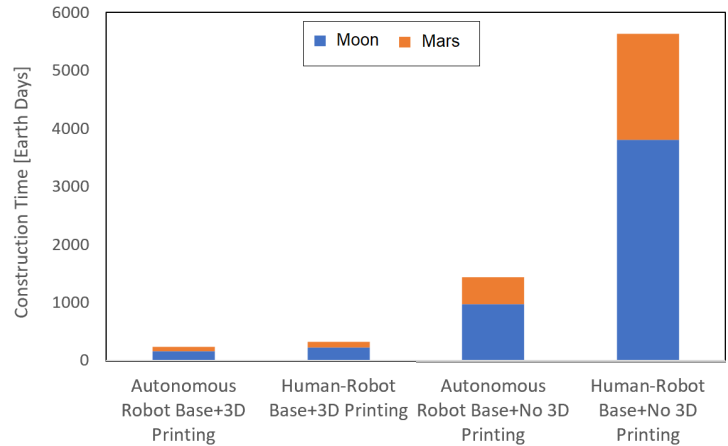

**Fig. 6. Comparison of Moon and Martian Mining Base Construction.** Using robotic and additive manufacturing can provide a 23-folds saving in construction time.

Based on our model, the significant simplifications possible of using an all-robot-team and housing robots is significantly more cost-effective than having humans and providing the significant resources needed to enable healthy living (Fig. 7, 8). If the resources needed for the base are obtained locally, that makes significant difference in terms of required transport energy and cost. Hence this study shows the potential game-changing opportunity possible with a construction method that utilizes local resources and doesn't require humans in the loop.

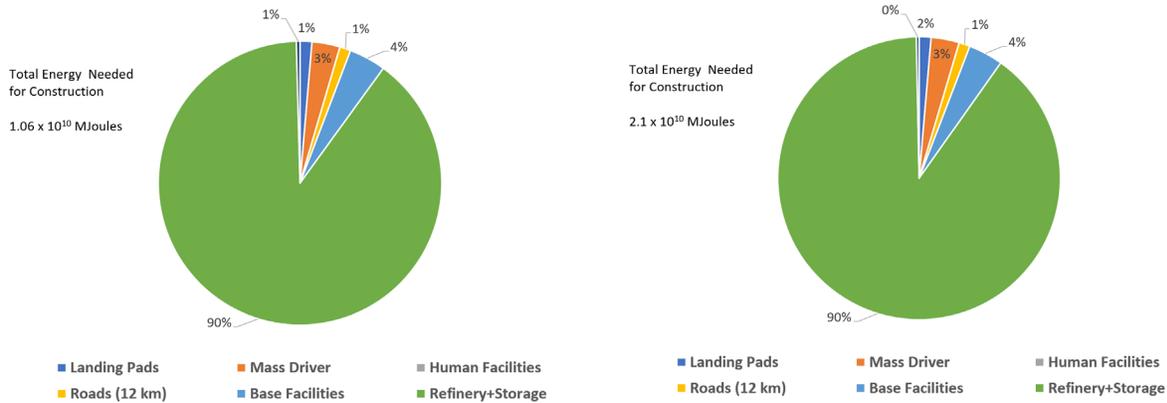

**Fig. 7. Distribution of energy consumed for building a robotic Moon (left) and Martian (right) mining base using 3D printing.**

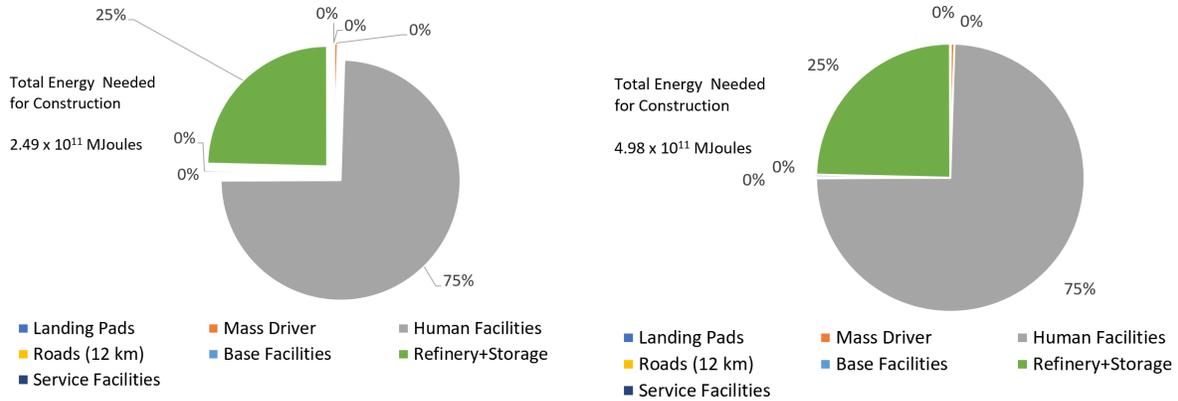

**Fig. 8. Distribution of energy consumed for building human occupied Moon (left) and Martian (right) mining base using conventional construction methods (non-3D printing).**

We look into the details of building the Off-World Robotic Mining Base (Fig. 7, 8). We presume 12 km of paved roads being built by fused silica-sand with properties similar to concrete. The roads will be reinforced with steel support beams to maximize strength for a 100-year lifetime. These roads will form the transport network for the fleet of 300 robot vehicles which is shown in Section IV.D. Importantly, the roads will connect the open-pit mine to the base and reflector facilities on the crater rim. The remainder will be base facilities including the power station, communication ground station and command and control facilities and 150-meter control tower.

The building structures are all domes that are 3D printed in place using the local silica sand and will be held with a reinforced steel structure that would also be locally produced from the iron/titanium rich regolith. The refineries will be the biggest dome structures on the base with a diameter of 100 m and hardened further to prevent any spill-over from damage due to accidents. The refinery infrastructure for the all-robot base will take up 90% of the construction energy. This is not surprising as the energy required for refining the metals dominant the daily energy consumption on the base (Fig. 9). The remaining 10% of the total energy will be spent on constructing human facilities, landing pads, roads, building the robot facilities including depot, maintenance and repair facilities, together with parts storage (Fig. 7).

**B. Energy Required for Base Maintenance**

According to our model, the total energy required for maintaining a Moon Mining base is $3.27 \times 10^7$ MJ per day. Of this, 86.6% of the required energy is for processing the 100 tons each of titanium, low-grade steel and aluminum. The energy needed to transport the 100 tons each of water, titanium, aluminum and steel into a lunar escape trajectory is 10.9 % of the total (Fig. 9). Remaining 2.5 % would be for mining, processing and maintaining the base facilities, while 0.24 % would be the energy required if there were to be 300 human workers. The significant amount of energy needed to refine and process the metal outweighs all other energy consumption needs include the energy needed to operate the Mass Driver. Building a Mass Driver [11] to transport the 100 tons each of water, titanium, steel and aluminum will be much easier than to do so from Earth or Mars due to low gravity of 1.62 m/s$^2$ (Fig. 9 right). Based on Fig. 10 (left), it takes 4.23-folds more energy to send the 600-ton payload mass into Mars escape as opposed to lunar escape velocity.

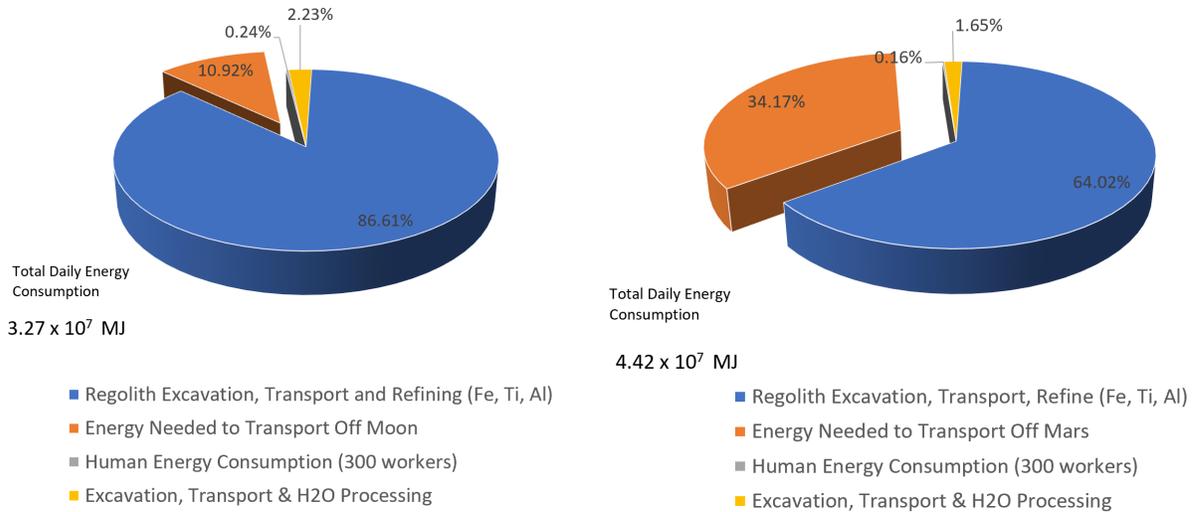

**Fig. 9. Distribution of energy consumed for operating a humans-occupied Moon (left) and Martian (right) Mining base.**

Comparing the energy required for resource extraction, nearly $2.9 \times 10^7$ MJ is consumed (Fig. 10 right). The extraction and processing of 100 tons each of titanium, low-grade steel and aluminum surrounding the base consumes 97.5% of this energy, while 2.5% of the energy is for extraction and transportation of water ice from the lunar south pole to the base located 1,000 km away via Maglev superconductive rail.

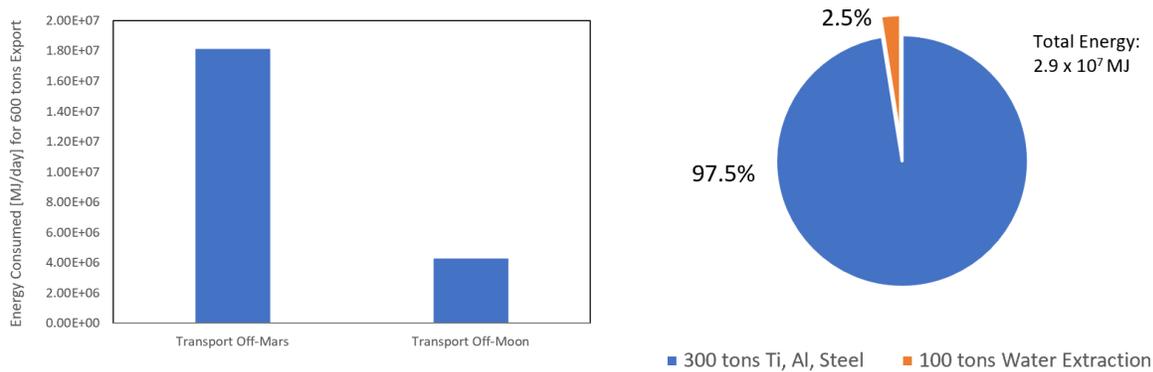

**Fig. 10. (Left) Energy Comparison for Mass Driver to transport 600 tons payload on escape velocity from the Mars and Moon. (Right) Distribution of energy consumed for extracting titanium, low-grade steel and aluminum in comparison to extraction of water-ice from the lunar south pole and transported 1000 km.**

**C. Human Impact**

Based on our models, the energy consumed per day (Earth day) for operation of the base (excluding Mass Driver transport and refining) is $8.02 \times 10^4$ MJ on the Moon and $7.23 \times 10^4$ MJ on Mars. Our studies show that having humans on the mining base results in significant consumption of energy (97-98%) and facility needs (Fig. 11).

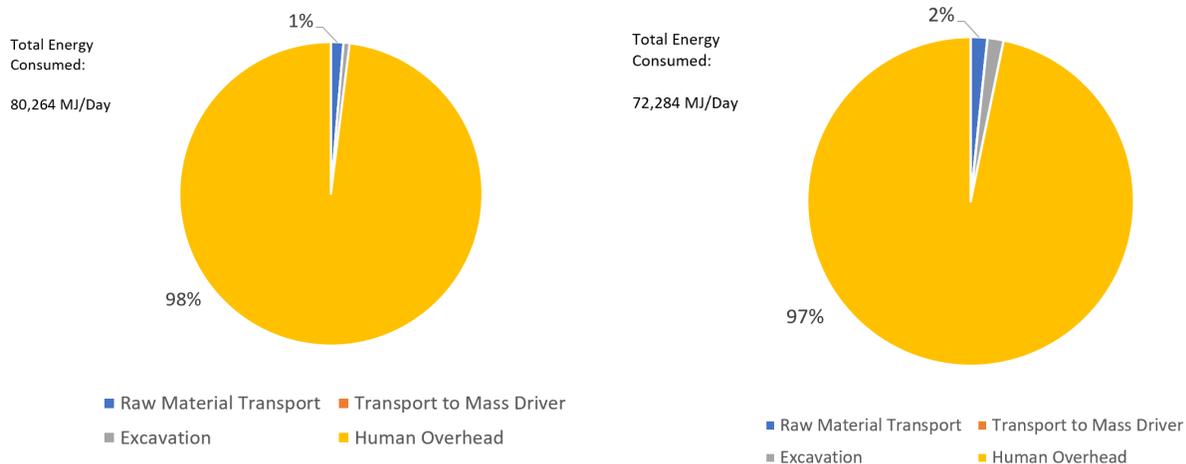

**Fig. 11. Distribution of energy consumed for operating a humans-occupied Moon (right) and Martian (left) Mining Base (excluding energy for Mass Driver and Refining).**

In fact, quite rapidly, human energy needs overtake all other energy needs except Mass Driver transport and refining of the raw material. This in part because Moon in its current form is an unfavorable environment to support Earth life and hence significant increase in energy and resources are required to sustain the life of the 300 human workers. In our energy model, we breakdown human needs into water, oxygen, food and electricity (Fig. 12). It is presumed each worker will require about 300 liters of water per day. The highest per capita usage of water is 550 liter per day in the United States. We presume 300 liters as it is attainable and will be possible with improved efficiency expected on a future lunar base.

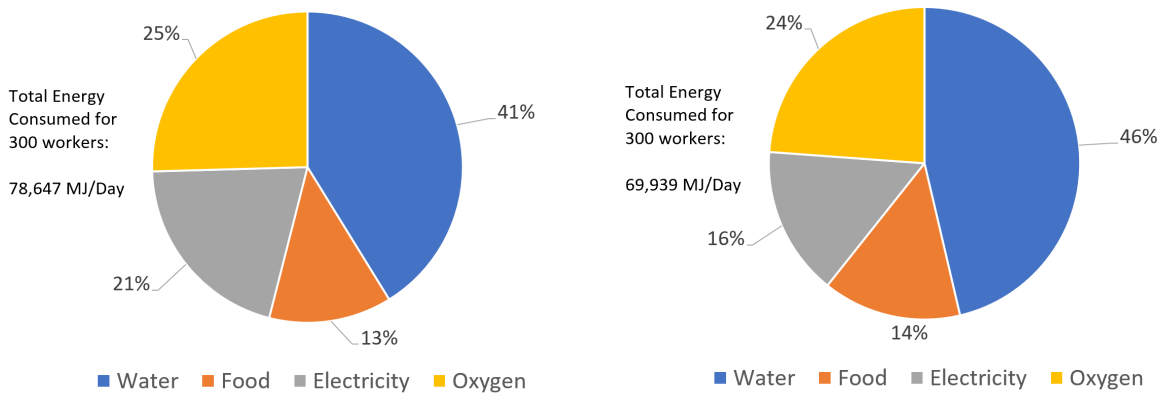

**Fig. 12. Distribution of energy consumed by 300 human workers on the Moon (right) and Martian (left) Mining Base.**

In addition, each worker is assumed to be consuming 15 kWh of electricity per day on the Moon and 10 kWh of electricity per day on Mars, in addition to daily consumption of 2,200 liters of oxygen. In terms of food (Fig. 12), each worker is estimated to consume nearly 1.8 kg of food (2,800 calories) per day, with 35% of the food being fats and proteins (including milk, eggs, meats and cheese), 17% being starch (wheat, corn, rice) and 48% (fruits, vegetables). Despite the difference in gravity we presume the workers will be consuming the same amount of food, with daily exercise routines to equalize for the different gravity conditions. We then accounted for net energy required to produce these foods. Beef is estimated to require 70.4 kJ/kg, while corn for example requires 1.1 kJ/kg. Using this detailed model, we find that the 300 workers consume 78,647 MJ/day on the Moon and 69,939 MJ/sol on Mars. This shows that automation and robotics is the key to making such a base technologically feasible. In the next section we provide a summary of our plans to further develop the 3D printing technology.

### D. Robotics Vehicles for 3D Printing, Excavation and Maintenance

The proposed robotic vehicles collect regolith and processes them onboard to perform 3D printing (Fig. 13, 14). The size of the 3D printed object is not limited by the size of the vehicle. The vehicles are powered entirely on renewable energy, using high-energy fuel cells [22-23] that provide double the energy output of gasoline. The robotic vehicles will also be autonomous operating as a group, with only high-level commands being provided by a human supervisor from Earth or a relay base [1-6] (see Fig. 15). Teams of autonomous vehicles operating under the right conditions can exceed human controllers.

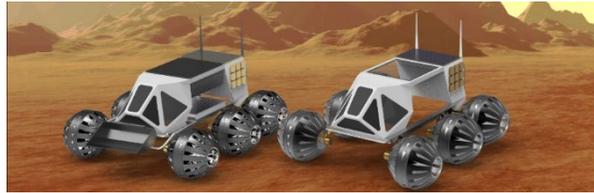

**Fig. 13. 3D Rendering of a pair of robotics vehicles for excavation and 3D printing.**

The vehicles are entirely autonomous and utilizes a combination of onboard sensors, GPS and local beacons to navigate. These vehicles are the template for a family of infrastructure robotic vehicles. On the moon they can be used on everything from digging trenches for underground conduits for water, air, power and data, to the construction and maintenance of pressurized greenhouses. They can be used to build roads, bridges, and more advanced infrastructures for human settlement including radiation-shielded architectures. Gravity being 16% that of Earth, smaller units can be effective at moving large volumes. The low gravity negatively impacts wheeled vehicles because it lowers their operating velocity before loosing traction or surface contact. There are artificial methods to increase rover traction by adding ballast weight. Furthermore, it has been found certain excavation implements are better suited for low-gravity than others. In particular our earlier studies found bucketwheel based excavators to have a productivity advantage and is easier to control in low-gravity environments [3].

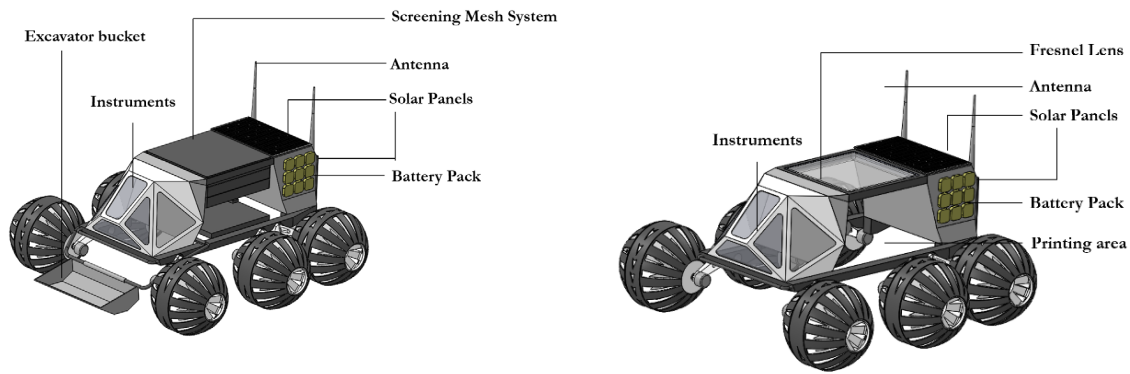

**Fig. 14. 3D Rendering of an autonomous robotic vehicle for excavation (left) and 3D printing (right).**

Based on these simulations, it is clear that base construction is significantly impacted by human occupation and lack of 3D printing. This has 23-fold increase in energy needed for based construction. The key factor is being able to perform site preparation, obtain raw material and perform 3D printing efficiently using robotics. Our work builds upon earlier work where we have used machine learning methods to evolve near-optimal controllers for multirobot systems to perform excavation, base preparation and open-pit resource mining (see Fig. 15) [1-6]. This shows promise and importance of efficient multirobot operation to cut-down on the energy needed for base construction which is the key, defining variable.

In comparison, other infrastructure elements such as the superconductive rail line consumes even more energy for construction. This is based on the presumption that large water reserves are only found in the south pole and where the temperatures are not conducive for a large mining/processing base. Thus, an efficient means of construction of the rail line is critical to making this entire enterprise feasible.

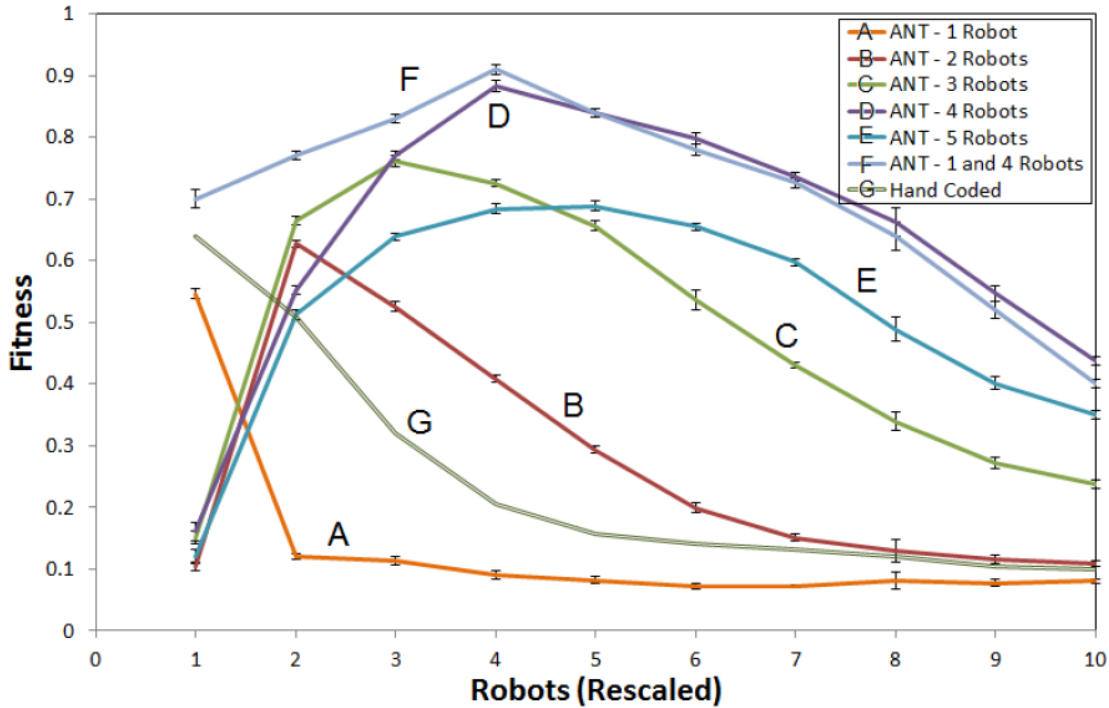

Fig. 15. Our earlier work shows that teams of robots for excavating task can be more effectively controlled using autonomous decentralized neuromorphic controllers than human-coded program [2, 12].

## V.   Conclusion

In this paper we modelled the design and construction of mining bases on the surface of the Moon and Mars. We presumed these mining base operate using renewable solar photovoltaic and solar-thermal energy.  We then compared construction techniques for the base including using conventional modular construction methods and additive manufacturing (or 3D printing).  We also compared the use of robot teams to construct and operate the base as an alternative to human worker teams.  Our study shows that the use of robotics to perform construction and operation is game-changing technology that can significantly speed-up operation of the base by 23-folds.  In fact, we find the benefit increases with increased base complexity.  A base that mines and exports various metals and water will obtain greater benefit from robotics and 3D printing than a base that just exports water. Our studies found that the use of human workers can significantly increase the complexity of base construction.  However, when it comes to day to day operations, human energy consumption is still limited compared to the energy cost of refining and transportation off-world.

Our findings show that there is significant energy savings possible from setting up and operating mining bases on the Moon rather than Mars due to the smaller-gravity well and relatively short distance from Earth to the Moon.  The only major hurdle to lunar mining is the need to build large infrastructure to extract water and that would heighten the capital cost of operations. In particular, high-speed Maglev cargo rail-lines have been identified for extracting and transporting water ice from the lunar PSR regions to mid-latitude region for further processing and export. Once constructed, the operational energy cost of transporting raw water ice using levitation technology has significant savings and is well coupled to the cold temperatures. Construction of this rail-line could well be facilitated through use of multiple robots to perform site preparation and construction using 3D printing techniques in-situ. Overall, we have identified a development pathway to utilize teams of robots to design, construct and operate complex mining bases off-world.

## Acknowledgments

This project received seed funding from the Dubai Future Foundation through Guaana.com open research platform.